\documentclass[10pt,twocolumn,letterpaper]{article}

\usepackage[utf8]{inputenc}
\usepackage{iccv}
\usepackage{times}
\usepackage{epsfig}
\usepackage{graphicx}
\usepackage{amsmath}
\usepackage{amssymb}
\usepackage{subfig}
\usepackage{url}
\usepackage{multirow}
\usepackage{booktabs}

\usepackage[breaklinks=true,bookmarks=false]{hyperref}

\iccvfinalcopy % *** Uncomment this line for the final submission

\ificcvfinal\fi

\begin{document}

%%%%%%%%% TITLE
\title{W-Net: A Facial Feature-Guided Face Super-Resolution Network}

\author{Hao Liu$^1$ \\
{\tt\small haoliu@mail.sdu.edu.cn}
\and
Yang Yang$^{1}$\thanks{Corresponding author.} \\
{\tt\small yyang@sdu.edu.cn}
\and
Yunxia Liu$^2$ \\
{\tt\small eyxliu@sdu.edu.cn}
\and
$^1$School of Information Science and Engineering, Shandong University, Qingdao, China\\
$^2$Center for Optics Research and Engineering (CORE), Shandong University, Qingdao, China
}

\maketitle
\ificcvfinal\thispagestyle{empty}\fi

%%%%%%%%% ABSTRACT
\begin{abstract}
Face Super-Resolution (FSR) aims to recover high-resolution (HR) face images from low-resolution (LR) ones. Despite the progress made by convolutional neural networks in FSR, the results of existing approaches are not ideal due to their low reconstruction efficiency and insufficient utilization of prior information. Considering that faces are highly structured objects, effectively leveraging facial priors to improve FSR results is a worthwhile endeavor. This paper proposes a novel network architecture called W-Net to address this challenge. 
W-Net leverages meticulously designed Parsing Block to fully exploit the resolution potential of LR image.  We use this parsing map as an attention prior, effectively integrating information from both the parsing map and LR images.  Simultaneously, we perform multiple fusions in various dimensions through the W-shaped network structure combined with the LPF(\textbf{L}R-\textbf{P}arsing Map \textbf{F}usion Module).  Additionally, we utilize a facial parsing graph as a mask, assigning different weights and loss functions to key facial areas to balance the performance of our reconstructed facial images between perceptual quality and pixel accuracy.  We conducted extensive comparative experiments, not only limited to conventional facial super-resolution metrics but also extending to downstream tasks such as facial recognition and facial keypoint detection.  The experiments demonstrate that W-Net exhibits outstanding performance in quantitative metrics, visual quality, and downstream tasks.

\end{abstract}

\begin{figure*}[htb!]
    \centering
    \includegraphics[width=1\linewidth]{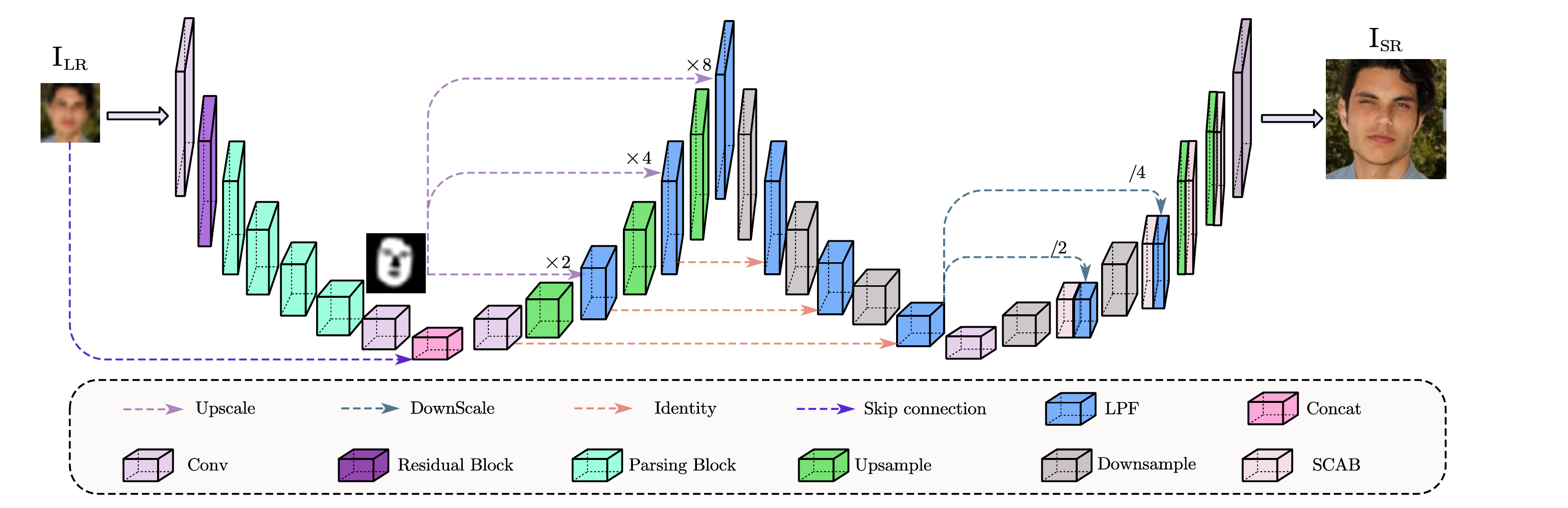}
    \caption{The W-Net model utilizes low-quality images to obtain face parsing maps as attention priors, effectively performing face super-resolution through the fusion of features from multiple scales of parsing maps and low-quality images.}
    \label{fig:overview}
\end{figure*}

%%%%%%%%% BODY TEXT
\section{Introduction}
Face Super-Resolution (FSR), also known as face hallucination, aims to reconstruct high-resolution (HR) images from low-resolution (LR) face images. In practical applications, the limitations of imaging devices often result in face images with reduced clarity, posing challenges for computer vision tasks such as face recognition\cite{Gao_2022} and face attribute analysis\cite{zheng2019survey}. This technology faces significant challenges due to the complex, unpredictable, and varied nature of image degradation.

A natural solution is to use a conventional Convolutional Neural Network (CNN) to directly learn the mapping from LR to HR images. Convolutional neural networks possess strong local modeling capabilities, allowing them to predict fine-grained facial details effectively. Zhou et al.\cite{zhou2015learning} developed the first FSR method based on CNNs. Recently, many FSR networks\cite{kim2019progressive,inproceedings,Zhang_2018_ECCV} have also been constructed using CNNs. However, these methods do not consider the limited capacity of deep learning frameworks and the prior knowledge inherent in facial structures.

As a special type of image object, the structured features of faces, such as facial heatmaps and facial parsing maps, provide rich prior information crucial for enhancing the performance of super-resolution techniques. In existing research, scholars have proposed various methods to leverage this prior information to guide the reconstruction process, aiming for better super-resolution results. FSRNet\cite{FSRNet} is the first end-to-end deep face super-resolution network that utilizes facial geometric priors. It consists of a coarse SR network, a fine SR encoder, a prior estimation network, and a final fine SR decoder. Based on the coarse SR network, facial landmarks and parsing maps are generated, and this prior knowledge is then used to complete the FSR task. Yu et al.\cite{Yu24} first generate intermediate features, then estimate facial heatmaps based on these features, and finally fuse the heatmaps with the intermediate features for the FSR task. However, both FSRNet and other methods use coarse networks to reconstruct low-resolution images, leading to an error accumulation phenomenon; if intermediate results are erroneous, the generated prior information and the final reconstructed results will also be flawed. Additionally, most of these methods only consider single-scale, one-time fusion, failing to fully utilize the prior information and low-quality image data for reconstruction.

The attention mechanism, inspired by the human visual system, has been integrated with deep learning FSR methods by many scholars\cite{Yu24,RCAN,SPARNet}. It is used to re-weight features to achieve the desired output, effectively assigning higher weights to the most informative convolutional features. However, relying solely on the attention mechanism may overlook other structural features of the face, which contain attribute and contextual information crucial for the reconstruction process. This can lead to suboptimal face images.

Considering the above problems, we design a W-shaped network, and in order to avoid error accumulation we simplify the parsing map problem and develop a Parsing Block that can fully exploit the model's ability to obtain face parsing information from low-quality maps. In addition, in order to avoid the defect of not being able to fully utilize the information of the parsed map, we not only design a LPF module(\textbf{L}R-\textbf{P}arsing Map \textbf{F}usion Module), but also perform multiple up-sampling and down-sampling operations on the parsed map, and perform multiple fusions in multiple dimensions.

In addition, to balance the perceptual quality and pixel accuracy, and to take into account the importance of key facial parts, we use simplified facial prior information as a mask to construct a new loss function. For eyes, eyebrows, nose and mouth, we assign special weights in the loss function to ensure better visual quality of the reconstructed facial images.

% Furthermore, in order to ensure global features in addition to the parsing map, this paper proposes SCAB by incorporating the convolutional self-attention module into this model based on the excellent work of RCAN\cite{RCAN}. To balance perceptual quality and pixel accuracy, and considering the importance of key facial parts, we used the simplified facial prior information as a mask to construct a new loss function. For the eyes, eyebrows, nose, and mouth, we assigned special weights in the loss function to ensure that the reconstructed facial images have better visual quality.

In summary, our key contributions are as follows:

1)We propose a W-shaped network architecture that integrates face parsing map estimation and face super-resolution processes into a unified framework. This design allows for the direct extraction of face parsing maps from LR images, which then guide the upscaling process to HR images, ensuring that the reconstructed faces retain accurate facial attributes and details.

2)We developed an LPF module that effectively combines complementary information from face parsing maps and LR image features. Using this module, we designed a W-shaped network architecture capable of performing multiple reconstructions at various scales, resulting in more robust and detailed HR face images.

3)To enhance the extraction of facial parsing information, we simplified the face parsing map into a binary matrix (0-1), where skin areas are represented by 1 and facial components and other parts by 0. Additionally, we developed a Parsing Block that integrates channel attention and spatial attention mechanisms, along with various feature extractors. This approach effectively explores the capability to transform low-quality images into accurate parsing maps.

4)We introduced a novel face parsing map-based loss function that allows our model to focus on different facial regions with varying importance. This loss function is tailored to the specific characteristics of facial features, ensuring high-quality reconstruction of key areas such as the eyes, eyebrows, nose, and mouth while maintaining the overall natural appearance of the face.

Our method has been extensively evaluated on standard benchmarks, demonstrating its effectiveness in generating HR face images that are not only higher in resolution but also more accurate and visually pleasing. Using face parsing information as a guiding prior has proven to be a powerful tool for improving FSR quality, offering a promising direction for future research and applications in this field.

%原来的Pasing Block
\begin{figure*}[ht]
\centering
\includegraphics[width=1\linewidth]{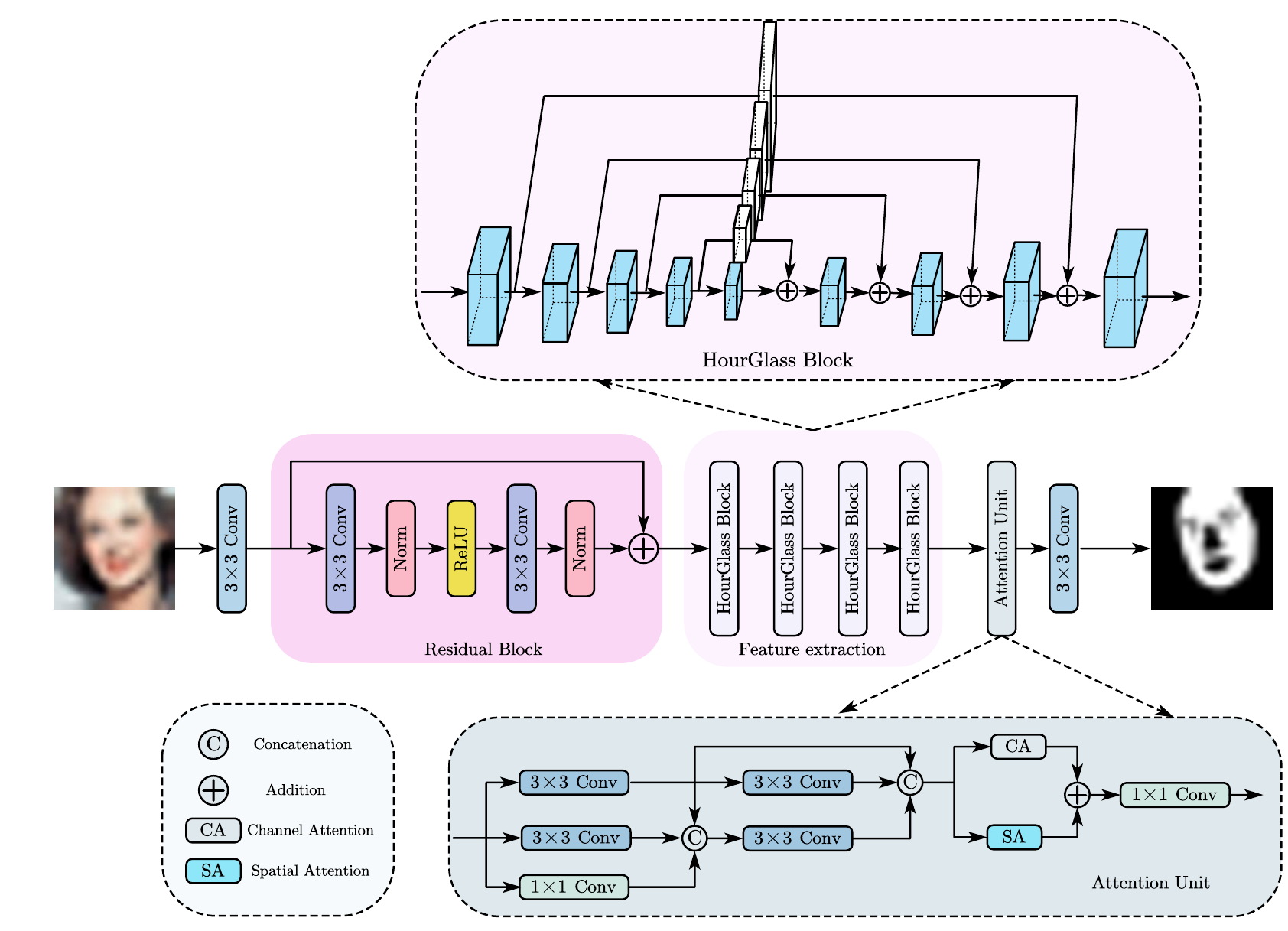}
\caption{\label{fig:Parsing Block}
The Parsing Block consists of shallow convolutional layers and residual blocks to extract deep features, followed by HourGlass blocks to extract facial landmark features. After passing through attention units and convolutional layers to adjust channel numbers, the facial parsing map is obtained.}
\end{figure*}

\section{Related Work}
Deep learning, a powerful machine learning technology, has demonstrated significant potential and achieved remarkable success in the field of face super-resolution (FSR). In this paper, we provide a comprehensive analysis and summary of deep learning-based FSR research, categorizing the work according to various foundational models and implementation approaches.
\subsection{Face Super-Resolution}
In 2000, Baker and Kanade\cite{Baker} first proposed the concept of face super-resolution. They adopted a method that searched for similar structures from training data to enhance the quality of low-resolution images. Since then, interest in face super-resolution technology has grown, leading to a series of related studies. Subsequently, Dong et al.\cite{dong2015image} introduced Convolutional Neural Networks (CNNs) into the field of image super-resolution. Due to the strong feature representation capabilities of deep CNN models, research on using deep CNNs for FSR has gradually increased, continuously improving performance.

Initially, deep learning-based FSR methods focused on designing efficient network architectures. For example, inspired by ResNet\cite{ResNet} and DenseNet\cite{DenseNet}, SRResNet\cite{SRGAN} and SRDenseNet\cite{SRDenseNet} were proposed to optimize network structures. To reduce computational complexity, Zhang et al.\cite{zhang} constructed the high-resolution information needed for reconstruction and recognition directly within the low-dimensional feature space.

Then Huang et al.\cite{WaveNet} found that wavelet transform could describe the texture and contextual information of images, so they performed face super-resolution in the wavelet domain and proposed WSRNet. Yu et al.\cite{URDGN} drew inspiration from Generative Adversarial Networks (GAN), which are known for their outstanding performance in image generation tasks. They innovatively applied the GAN concept to the FSR field and proposed URDGN. To make the generated facial images more realistic and detailed, AGA-GAN\cite{AGAGAN} is proposed, which employs a new attribute-guided attention module to identify and focus on the generation process of various facial features in the images.

However, the aforementioned methods did not consider high-frequency details, which are a significant challenge in image super-resolution. Therefore, HiFaceGAN\cite{HifaceGAN} designed an effective suppression module specifically for recovering high-frequency details in images. Considering that the faces in low-quality and high-quality images might be misaligned, Yu et al.\cite{TransformYu} integrated a spatial transformation network\cite{TransformNetwork} into the CNN architecture to align LR face images.

Although CNNs perform well in image processing, they may be limited in capturing long-range dependencies and global information. To overcome these limitations, attention mechanisms have become an important research direction. Zhang et al.\cite{RCAN} proposed a Residual Channel Attention Block (RCAN), which generates different attentions for each channel feature to improve the network's discriminative ability. Considering the restoration of more detailed local features, Chen et al.\cite{SPARNet} designed a facial attention mechanism. Lu et al.\cite{lu2021face} proposed an internal feature segmentation attention mechanism to better capture facial semantic information for face super-resolution tasks.

To better handle global information in images, attention-based architectures like Transformers\cite{Attention} have been receiving increasing attention. Their key feature is the self-attention mechanism, which effectively captures long-range correlations between words or pixels. Lu et al.\cite{Lu_2022_CVPR} simplified the Transformer structure by utilizing only the encoder for self-attention, proposing ELAN for SR and achieving competitive results. Subsequently, Liang et al.\cite{SwinIR}, supported by the Swin-Transformer\cite{SwinTransformer}, introduced SwinIR for image restoration, renowned for its effective feature extraction capabilities. While Transformer-based methods generally excel in SR tasks, their global mechanisms pose challenges in extracting local textures\cite{Peng_2021_ICCV}, which in turn affects the visual quality and facial features in face reconstruction.

% To better handle global information in images, Transformers\cite{Attention}, an attention-based architecture, have gradually gained attention. The key idea is the self-attention mechanism, which can capture long-range correlations between words/pixels. Lu et al. \cite{Lu_2022_CVPR} simplified the Transformer structure, using only the encoder for self-attention, and proposed ELAN for SR, achieving competitive results. Subsequently, supported by Swin-Transformer\cite{SwinTransformer}, Liang et al.\cite{SwinIR} proposed SwinIR for image restoration, which has effective feature extraction capabilities. Although Transformer-based methods perform well in general SR problems, the global mechanism makes it difficult to extract local textures\cite{Peng_2021_ICCV}, affecting the visual quality and facial features in face reconstruction.

The above work indicates that relying solely on CNNs or Transformers has certain limitations. To better improve reconstruction quality, CNN-Transformer network architectures have been proposed. Wang et al.\cite{TANet} designed TANet, which integrates CNN and Transformer, but it only simply connects the features of CNN and Transformer. Considering deeper combinations, Gao et al.\cite{CTCNet} proposed CTCNet, which effectively combines global information and local features of images for high-quality image reconstruction.

\subsection{Prior Based Method}
As highly structured objects, face images possess inherent structural knowledge or prior information that can enhance the effectiveness of face super-resolution (FSR). Consequently, prior-guided FSR methods have gained significant attention and achieved impressive milestones.

Firstly, geometric priors of faces, including facial heatmaps, facial landmarks, and facial parsing maps, have been widely applied in FSR tasks. Early on, LCGE\cite{LCGE} adopted a pre-trained landmark detection model to divide the entire face image into different components, each of which was restored by different models. However, for super-resolution tasks, the landmark information provided by low-resolution images is limited, and relying solely on pre-trained models is insufficient to extract this information. To address this, researchers have attempted to extract facial parsing maps and facial heatmaps. Yu et al.\cite{YuHeatMap} developed a convolutional neural network with two branches: one for estimating facial component heatmaps and the other for reconstructing facial images with the help of these heatmaps. Recognizing that multiple priors can aid in image restoration, Xiu et al.\cite{Xiu2022DoubleDF} adopted a two-branch approach to fully utilize both facial heatmap and landmark information.

To reduce the difficulty of estimating priors from low-resolution (LR) images, using iterative intermediate results has become a research hotspot. FSRNet\cite{FSRNet} first recovers a coarse super-resolution face image to enrich facial information, then uses the intermediate results to extract facial priors, thereby reducing the difficulty of prior estimation. The extracted priors and intermediate results are then merged to obtain the final high-quality face image. Similarly, Ma et al.\cite{DIC} developed DIC, which first performs super-resolution on LR face images to obtain super-resolution results. These results are then used to estimate priors, which are combined with intermediate results in subsequent iterations, continuously improving the FSR restoration results. Wu et al.\cite{WU2023104857} proposed a robust semantic prior-guided FSR framework for face reconstruction.

However, with iterative methods, the overly simple structure of shallow networks makes the intermediate results of facial reconstruction prone to inaccuracies and errors during iteration. This can lead to the accumulation and amplification of errors, ultimately deteriorating the subsequent face reconstruction.

% Geometric priors of faces have proven to be significant in FSR tasks. We focused on facial parsing maps, simplifying the problem to avoid the deterioration of face reconstruction. At the same time, we considered using multi-dimensional and multiple fusions to fully utilize the valuable facial parsing map information.

In addition to the aforementioned FSR method we introduced, there are many other approaches related to FSR. For example, the ASFFNet\cite{ASFFNet} which is based on multiple face reference priors, the DMDNet\cite{DMDNet} which is based on dictionaries, the CodeBook-based method\cite{CodeBook}, reinforcement learning-based methods\cite{cao2017attentionaware}, and knowledge distillation-based methods\cite{zhishizhengliu},among others.

\section{Methods}
For LR images produced by Bicubic interpolation, our W-Net aims to complete the super-resolution task by extracting face parsing information from low-quality images and using this prior information. The proposed face restoration model can be defined as follows:
\begin{equation}
\begin{aligned}
\hat{I}^h=\mathcal{F}(I^d|Parsing^L;\Theta),
\end{aligned}
\end{equation}
where $Parsing^L$ represents the face parsing information obtained directly from the low-quality image, and $\Theta$ represents the learnable parameters of the model. 

Our model framework is shown in Figure ~\ref{fig:overview}, which overall presents a "W" shape, with the face parsing map estimation and super-resolution process integrated into a unified framework. Below, we first introduce the simplified process of face parsing map estimation and the Parsing Block we proposed to address the parsing problem. Then, we introduce the LPF Module to fully integrate the parsing map information with the low-frequency information from the low-quality image. Finally, we present the training and learning objectives of the entire framework.

\begin{figure*}[ht]
\centering
\includegraphics[width=1\linewidth]{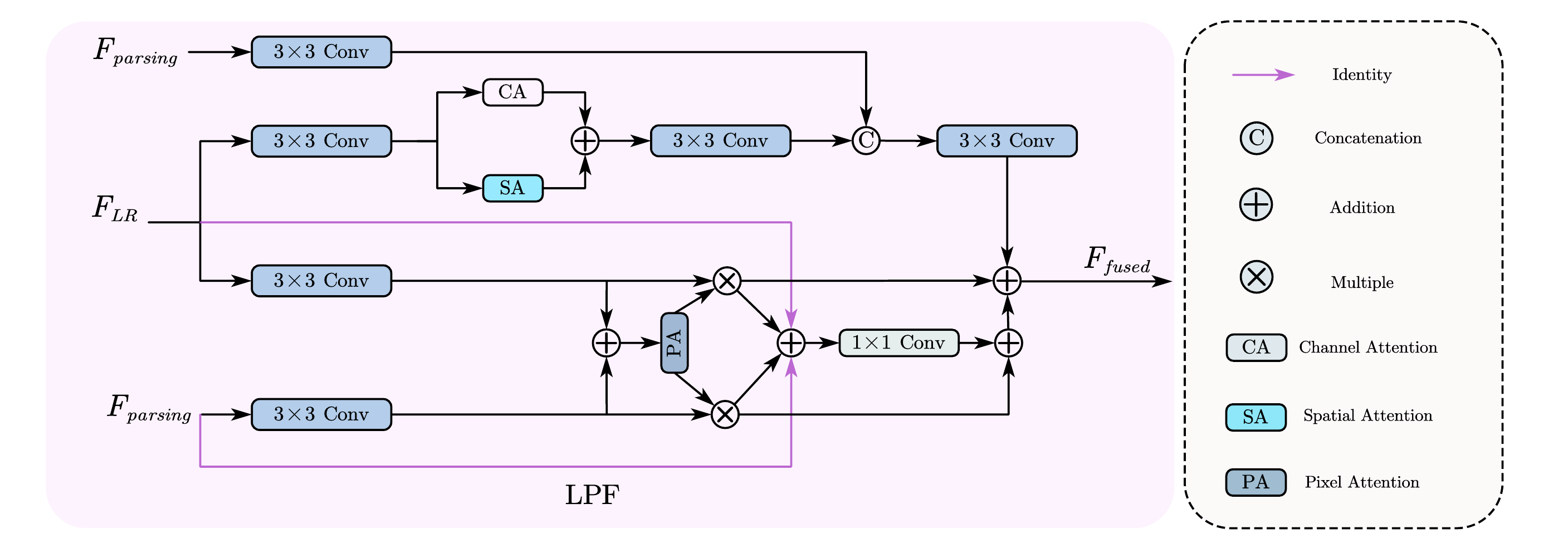}
\caption{\label{fig:Fuse Block}
The LPF is composed of multiple convolutional layers and different attention layers. It weights the LR and ParsingMap at the pixel level. Simultaneously, it utilizes multiple identity connections to form the final output.}
\end{figure*}

\subsection{Parsing problem simplification}
To simplify the parsing map problem, we binarize the parsing map obtained using a pre-trained BiseNet\cite{BiseNet}, setting the face region to 1 and the facial features and background to 0. The resulting face parsing map is thus a simple 0-1 matrix. With the support of a large amount of data and our proposed Parsing Block, this approach is sufficient to generate accurate parsing information from low-quality images.

Whether in low-quality face parsing tasks or in face super-resolution (FSR) tasks, the main challenge is extracting key facial features (such as eyes, eyebrows, nose, and mouth) and ensuring the network focuses on these features. To address this, we propose a Parsing Block to enable our model to extract as much useful information as possible, thereby improving detail recovery. Its structure is shown in Figure ~\ref{fig:Parsing Block}.

Since residual blocks and hourglass structures have been successfully used in human pose estimation and FSR tasks\cite{10.1109/TCSVT.2017.2778227,HourGlass}, we first use a convolution module to extract shallow features and expand the number of input channels. Next, we employ residual blocks combined with an hourglass structure to capture facial landmark features. Subsequently, an attention unit composed of spatial attention and channel attention enhances the representation capability of the extracted facial features. Finally, a face parsing map is generated to complete the super-resolution task.

Specifically, given an LR image, we first use $3\times3$ convolution to extract shallow features.

\begin{equation}
\begin{aligned}
{F}_{Shallow}{=}{f}_{3\times3,64}{(I}_{LR}{)},
\end{aligned}
\end{equation}
where ${f}_{3\times3,64}$ represents the convolution operation directly applied to the LR image using a convolution kernel of size $3 \times 3$ with 64 output channels.

Then, we further extract features from the shallow features using residual blocks and an HourGlass structure.
\begin{equation}
\begin{aligned}
F_{deep}=\mathcal{F}_{HG}\big(\mathcal{F}_{Res}\big(F_{Shallow}\big)\big),
\end{aligned}
\end{equation}
where $\mathcal{F}_{Res}$ represents the residual block using $3 \times 3$ convolution kernels and ReLU as the activation function, and $\mathcal{F}_{HG}$ represents the HourGlass module with a depth of 4 for feature extraction.

Then, we introduce a multi-scale feature attention unit. We use stacked convolution layers and attention mechanism layers to extract face parsing information from the features. First, we use a series of convolutions with different scales to obtain feature maps with different receptive fields, specifically including:

\begin{equation}
\begin{aligned}
y_1=\mathcal{F}_{act}\big(f_{1\times1}\big(F_{Deep}\big)\big),\\y_2=\mathcal{F}_{act}\big(f_{3\times3}\big(F_{Deep}\big)\big),\\y_3=\mathcal{F}_{act}\big(f_{3\times3}\big(F_{Deep}\big)\big),
\end{aligned}
\end{equation}
where, $y_1$ represents using a $1 \times 1$ convolution to adjust the channel size, and represent using $3 \times 3$ respectively to reduce the dimensionality of the feature maps.

Next, we concatenate the feature maps and process both the concatenated feature map and the original feature map separately.
\begin{equation}
\begin{aligned}y_4&=\mathcal{F}_{act}\big(f_{3\times3}\big(\varpi\big(y_1,y_2,y_3\big)\big)\big),\\y_5&=\mathcal{F}_{act}\big(f_{3\times3}\big(y_3\big)\big),\end{aligned}
\end{equation}
where $\varpi$ represents concatenating the three feature maps obtained from different convolutional kernels along the channel dimension. Then, we use a convolution layer with doubled channels to extract features from the concatenated feature map, and a convolution layer with halved channels to extract features from the original feature map to obtain more abstract features.

Next, we recombine the abstract features, concatenated features, and original features obtained, and incorporate attention mechanisms both spatially and across channels.
\begin{equation}
\begin{aligned}y_{CA}&=CA\left(\varpi\left(y_2,y_4,y_5\right)\right),\\y_{SA}&=SA\left(\varpi\left(y_2,y_4,y_5\right)\right),\\y_{{6}}&=y_{CA}+y_{SA},\end{aligned}
\end{equation}
where CA represents using channel attention to process the concatenated feature map, enhancing the relationships between feature maps and improving feature representation. SA represents using spatial attention to process the concatenated feature map, capturing local information in the image.

We adjust the channel size of the feature map that contains both local and feature information using $1 \times 1$ convolutional layers, and then use residual connections with the initial feature map to obtain the final output.
\begin{equation}
\begin{aligned}
y_7 &= f_{1\times1}\begin{pmatrix}y_6\end{pmatrix}, \\
y_{feature~out} &= y_7 + F_{Deep}.
\end{aligned}
\end{equation}

Finally, to obtain the final parsing map, we use another convolutional layer with 3 output channels to adjust the channel size.
\begin{equation}
\begin{aligned}y_{out}=f_{3\times3,3}\left(y_{\textit{feature out}}\right).\end{aligned}
\end{equation}

\subsection{LR-Parsing Map Fusion Module(LPF)}
As one of the most important modules in W-Net, LPF is designed to integrate the features of the high-resolution image and the low-quality image. It not only preserves the feature information of the high-resolution image but also incorporates the contextual information of the low-quality image. As shown in Figure ~\ref{fig:Fuse Block}, we use different convolutional kernels to process the feature maps. Subsequently, we concatenate and additively combine the feature maps generated from the low-quality image and the high-resolution image. Particularly, for the additively combined feature maps, we employ a weighted summation approach for integration. Additionally, we introduce two skip connections to effectively prevent the vanishing gradient issue by connecting the output with the initial feature map and the feature map after convolution.

Specifically, three convolutional kernels are utilized to process the given low-quality feature map $F_{LR}$ and the high-resolution feature map separately$F_{Parsing}$.
\begin{equation}
\begin{aligned}\begin{gathered}
{F}_{LR}^{(1)} \Large=f_{3\times3}(F_{LR}), \\
{F}_{LR}^{(2)} \Large=f_{3\times3}(F_{LR}), \\
{F}'_{parsing} \Large=f_{3\times3}(F_{parsing}).
\end{gathered}\end{aligned}
\end{equation}

After receiving the features extracted by convolution, we will employ parallel processing. On one side, we apply channel attention mechanism and spatial attention mechanism separately, and then sum up the results of these two processes. On the other side, we directly sum up the features after convolution, and apply pixel-level attention mechanism for further processing to obtain feature adjustment weights. These weights are further combined to obtain the fused features.

\begin{equation}
\begin{aligned}
\begin{aligned}
F_{\mathrm{Att}}& =f_{3\times3}\left(CA\left(F_{LR}^{(1)}\right)+SPA\left(F_{LR}^{(1)}\right)\right),  \\
F_{\mathrm{pixel}}& =\mathrm{~}\sigma\cdot F_{LR}^{(2)}+(1-\sigma)\cdot F_{parsing}^{\prime},  \\
F_{\mathrm{fuse}1}& =f_{3\times3}(\varpi(F_{\mathrm{Att}},F_{parsing}^{\prime})),  \\
F_{\mathrm{fuse}2}& =f_{1\times1}(F_{\mathrm{pixel}}+F_{LR}+F_{parsing}), 
\end{aligned}
\end{aligned}
\end{equation}
where $\sigma=PA\left(F_{LR}^{(2)}+F_{parsing}^{\prime}\right)$ represents the feature map weights obtained using pixel-level attention.

% \begin{equation}
% \begin{aligned}
% F_{\text{fuse}1}=f_{3\times3}\big(\varpi\big(f_{3\times3}\big(CA\big(F_{LR}^{(1)}\big)+SPA\big(F_{LR}^{(1)}\big)\big),F'_{parsing}\big)\big)\\F_{\text{fuse}2}=f_{1\times1}\big(\sigma\cdot F_{LR}^{(2)}+\big(1-\sigma\big)\cdot F'_{parsing}+F_{LR}+F_{parsing}\big),
% \end{aligned}
% \end{equation}

The existing feature maps are further combined to produce the final output through skip connections.
\begin{equation}
\begin{aligned}F_{fused}=F_{\text{fuse}1}+F_{\text{fuse}2}+F_{LR}^{(2)}+F_{parsing}^{\prime}.\end{aligned}
\end{equation}

\begin{figure*}[htb!]
    \centering
    \includegraphics[width=1\linewidth]{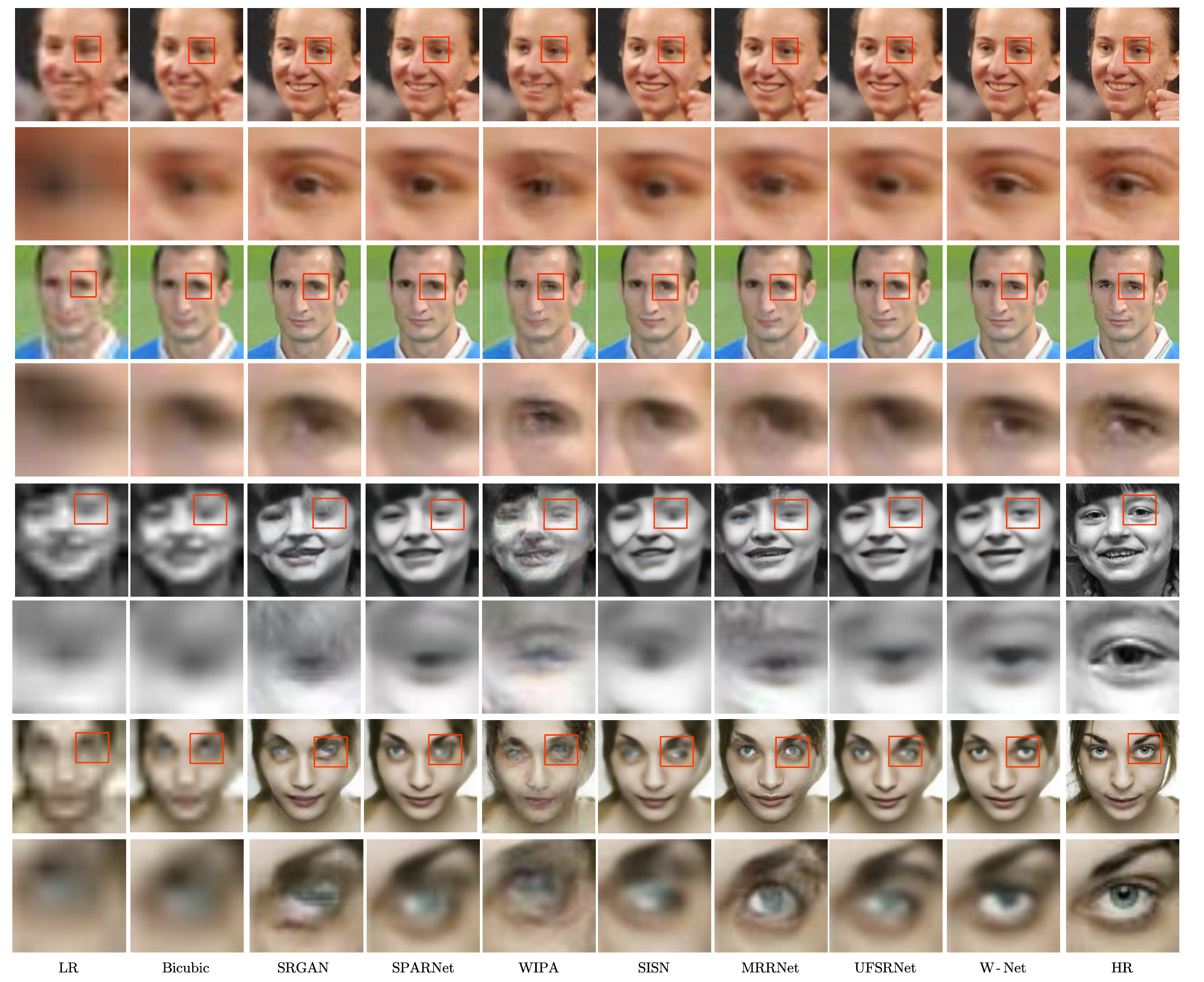}
    \caption{Visual comparison with state-of-the-art facial super-resolution methods. The low-resolution facial images are sized 32×32 (top two rows) and 16×16 (bottom two rows), upscaled by factors of four and eight, respectively. Better zoom in to see the detail.}
    \label{fig:Compare}
\end{figure*}

\begin{table*}[ht]
    \centering
    \scriptsize
    \caption{Quantitative comparison of the various FSR methods on the Helen and CelebA datasets. The results for Training Mode I are shown above and those for Training Mode II are shown below. The best and second best results are bolded and underlined, respectively.}
    \label{tab:Compare}
    \begin{tabular}{lccccccccccccc}
        \toprule
        \textbf{Methods} & \textbf{Year} & \multicolumn{3}{c}{CelebA($ \times 4 $)} & \multicolumn{3}{c}{CelebA($ \times 8 $)} & \multicolumn{3}{c}{Helen($ \times 4 $)} & \multicolumn{3}{c}{Helen($ \times 8 $)} \\
        \cmidrule(lr){3-5} \cmidrule(lr){6-8} \cmidrule(lr){9-11} \cmidrule(lr){12-14}
         & & \textbf{PSNR$\uparrow$} & \textbf{SSIM$\uparrow$} & \textbf{LPIPS$\downarrow$} & \textbf{PSNR$\uparrow$} & \textbf{SSIM$\uparrow$} & \textbf{LPIPS$\downarrow$} & \textbf{PSNR$\uparrow$} & \textbf{SSIM$\uparrow$} & \textbf{LPIPS$\downarrow$} & \textbf{PSNR$\uparrow$} & \textbf{SSIM$\uparrow$} & \textbf{LPIPS$\downarrow$} \\
        \midrule
        Bicubic & - & 27.38 & 0.8002 & 0.1857 & 23.46 & 0.6776 & 0.2699 & 28.12 & 0.8423 & 0.1771 & 23.80 & 0.6396 & 0.2560 \\
        SRCNN\cite{SRCNN} & 2014 & 28.01 & 0.8332 & 0.1489 & 24.01 & 0.6747 & 0.2559 & 28.69 & 0.8715 & 0.0558 & 24.31 & 0.6780 & 0.2471 \\
        EDSR \cite{EDSR} & 2017 & 31.50 & 0.9001 & 0.0513 & 26.99 & 0.7790 & 0.1144 & 31.85 & {0.9128} & 0.0579 & 26.57 & 0.7843 & 0.1442 \\
        FSRNet\cite{FSRNet} & 2018 & 31.37 & 0.9012 & 0.0501 & 26.86 & 0.7714 & 0.1098 & 31.97 & 0.9188 & 0.0553 & 26.49 & 0.7802 & 0.1382 \\
        DIC\cite{DIC} & 2020 & \underline{31.58} & \underline{0.9015} & 0.0532 & \underline{27.35} & \underline{0.8019} & 0.0902 & \underline{32.01} & \underline{0.9223} & 0.0587 & \underline{26.98} & \underline{0.8015} & 0.1158 \\
        MRRNet\cite{MRRNet} & 2022 & 30.20 & 0.8687 & \textbf{0.0278} & 26.09 & 0.7430 & \textbf{0.0592} & 31.10 & 0.9013 & \textbf{0.0328} & 26.54 & 0.7810 & \textbf{0.0619} \\
        \textbf{W-Net} & 2024 & \textbf{31.77} & \textbf{0.9032} & \underline{0.0482} & \textbf{27.54} & \textbf{0.8041} & \underline{0.0908} & \textbf{32.32} & \textbf{0.9251} & \underline{0.0491} & \textbf{27.26} & \textbf{0.8121} & \underline{0.1046} \\
        \midrule
        SRGAN\cite{SRGAN} & 2017 & 31.05 & 0.8880 & 0.0459 & 26.63 & 0.7628 & 0.1043 & 31.01 & 0.9002 & 0.0499 & 25.83 & 0.7491 & 0.1109 \\
        SPARNet\cite{SPARNet} & 2020 & 31.52 & 0.9005 & 0.0593 & \underline{27.29} & \underline{0.7965} & 0.1088 & 31.72 & \underline{0.9171} & 0.0682 & 26.95 & 0.8029 & 0.1169 \\
        SISN\cite{lu2021face} & 2021 & \underline{31.55} & \underline{0.9010} & 0.0587 & 26.83 & 0.7786 & {0.1044} & \underline{31.73} & 0.9163 & 0.0708 & 26.07 & 0.7680 & {0.1305} \\
        WIPA\cite{WIPA} & 2022 & 30.35 & 0.8711 & 0.0619 & 26.23 & 0.7652 & 0.0961 & 30.47 & 0.8923 & 0.0738 & 26.75 & 0.7514 & {0.1202} \\
        MRRNet\cite{MRRNet} & 2022 & 30.48 & 0.8720 & \textbf{0.0374} & 25.94 & 0.7417 & \textbf{0.0562} & 30.80 & 0.8951 & \textbf{0.0456} & 26.20 & 0.7731 & \textbf{0.0661} \\
        UFSRNet \cite{UFSRNet} & 2024 & 31.42 & 0.8987 & 0.0643 & 27.10 & 0.7887 & 0.0791 & 31.64 & {0.9159} & 0.0704 & \underline{26.99} & \underline{0.8031} & 0.1218 \\
        \textbf{W-Net} & 2024 & \textbf{31.63} & \textbf{0.9029} & \underline{0.0425} & \textbf{27.40} & \textbf{0.8014} & \underline{0.0760} & \textbf{31.83} & \textbf{0.9181} & \underline{0.0483} & \textbf{27.05} & \textbf{0.8058} & \underline{0.1028} \\
        \bottomrule
    \end{tabular}
\end{table*}

\begin{table*}[htb]
\centering
\caption{In Training Mode II, the quantitative evaluation of various FSR models on downstream tasks uses mean Euclidean distance and mean identity cosine similarity as comparison indices.}
\label{tab:xiayou}
\resizebox{\textwidth}{!}{%
\begin{minipage}{\textwidth}
\centering
\footnotesize
\begin{tabular}{c@{\hskip 0.05in}c@{\hskip 0.05in}c@{\hskip 0.05in}cccccc} % 表格列声明
\toprule
Dataset & Scale & Metric & Bicubic & SPARNet\cite{SPARNet} & SRGAN\cite{SRGAN} & MRRNet\cite{MRRNet} & UFSRNet\cite{UFSRNet} & \textbf{W-Net} \\ \midrule
\multirow{3}{*}{CelebA} & \multirow{2}{*}{$\times4$} & Euclidean distance$\downarrow$ & 11.06 & \underline{7.58} & 7.65 & 7.72 & 7.59 & \textbf{6.98} \\
 & & Cosine similarity$\uparrow$ & 0.7339 & \underline{0.8413} & 0.8306 & 0.8267 & 0.8371 & \textbf{0.8697} \\
\cmidrule{2-9}
 & \multirow{2}{*}{$\times8$} & Euclidean distance$\downarrow$ & 22.23 & \underline{11.12} & 11.72 & 12.15 & 11.96 & \textbf{10.17} \\
 & & Cosine similarity$\uparrow$ & 0.3680 & 0.5744 & 0.5465 & \underline{0.5795} & 0.5530 & \textbf{0.5861} \\ \midrule
\multirow{3}{*}{Helen} & \multirow{2}{*}{$\times4$} & Euclidean distance$\downarrow$ & 11.74 & 9.07 & \underline{9.00} & 9.22 & 9.08 & \textbf{8.95} \\
 & & Cosine similarity$\uparrow$ & 0.7767 & \underline{0.8612} & 0.8516 & 0.8660 & 0.8683 & \textbf{0.8690} \\
\cmidrule{2-9}
 & \multirow{2}{*}{$\times8$} & Euclidean distance$\downarrow$ & 24.15 & 13.91 & 16.43 & 14.30 & \underline{13.34} & \textbf{13.28} \\
 & & Cosine similarity$\uparrow$ & 0.3858 & 0.6483 & 0.5584 & \textbf{0.6810} & 0.6360 & \underline{0.6545} \\
\bottomrule
\end{tabular}
\end{minipage}
}
\end{table*}

\subsection{Multi-scale multiple reconstruction network}
To effectively integrate both the geometric structural information from the face parsing map and the contextual information from the low-quality image for super-resolution tasks, considering that single-scale fusion may miss a lot of information, we adopt a strategy of upscale-merge-upscale-merge. This involves multiple rounds of fusion, where the features are further extracted through downsampling and then reconstructed into upsampled images.

Specifically, given the LR and the  generated parsing map , we first concatenate them along the channel dimension. This constitutes the \textbf{initial} direct fusion step.
\begin{equation}
\begin{aligned}{F}_{fused}^{(1)}={\varpi}({I}_{LR},{Parsing}^L),\end{aligned}
\end{equation}
where ${Parsing}^L$ is the parsing map corresponding to the LR.

Then, we upsample ${Parsing}^L$ by  factors of 2 , 4 and 8 using nearest-neighbor interpolation.
\begin{equation}
\begin{aligned}\begin{gathered}
{I}_{\mathrm{NN}}^1 \Large=NN(Parsing^L,1), \\
{I}_{\mathrm{NN}}^2 \Large=NN(Parsing^L,2), \\
{I}_{\mathrm{NN}}^4 \Large=NN(Parsing^L,4), \\
{I}_{\mathrm{NN}}^8 \Large=NN(Parsing^L,8), 
\end{gathered}\end{aligned}
\end{equation}
where $\mathrm{NN}(\cdot)$ represents nearest-neighbor interpolation for upsampling operation.

Next, We apply a $3 \times 3$ convolution operation to the fused feature map obtained after the first fusion to extract features, adjusting the channel number to 64. Then, we use an upsampling module to upscale this combined feature, resulting in a feature map enlarged by a factor of 2.
\begin{equation}
\begin{aligned}F_{up}^{(2)}{=}Upsample\left(\begin{array}{c}f_{3\times3,64}(F_{fused}^{(1)})\end{array}\right),\end{aligned}
\end{equation}
where $F_{up}^{(2)}$ represents the features enlarged by a factor of 2, and $Upsample(\cdot)$ represents the upsampling module, consisting of Pixel Shuffle layers and BatchNorm layers.

Then, we use the LPF to perform the \textbf{second} feature fusion on the features enlarged by a factor of 2 and the parsing map enlarged by a factor of 2.
\begin{equation}
\begin{aligned}{F}_{fused}^{(2)}{=}{LPF}({F}_{up}^{(2)}{,I}_{\mathrm{NN}}^2).\end{aligned}
\end{equation}

Next, we repeat the above operations: we further upscale ${F}_{fused}^{(2)}$ to obtain features enlarged by a factor of 4 and perform the \textbf{third} fusion with the parsing map enlarged by a factor of 4. 
\begin{equation}
\begin{aligned}
F_{up}^{(4)}&=Upsample\Big(f_{3\times3,64}\Big(F_{fused}^{(2)}\Big)\Big),\\F_{fused}^{(3)}&=LPF\big(F_{up}^{(4)},I_{\mathrm{NN}}^{4}\big).
\end{aligned}
\end{equation}

Similarly, we upscale ${F}_{fused}^{(3)}$ to obtain features enlarged by a factor of 8 and perform the \textbf{fourth} fusion with the parsing map enlarged by a factor of 8.
\begin{equation}
\begin{aligned}
F_{up}^{(8)}&=Upsample\Big(f_{3\times3,64}\Big(F_{fused}^{(3)}\Big)\Big),
\\F_{fused}^{(4)}&=LPF\big(F_{up}^{(8)},I_{\mathrm{NN}}^{8}\big).
\end{aligned}
\end{equation}

At this point, the size of the feature map is $1024 \times 1024$. Next, we downsample it and further perform fusion.
\begin{equation}
\begin{aligned}F_{down}^{(4)}&=Downsample\big(f_{3\times3,64}\big(F_{fused}^{(4)}\big)\big),\\F_{fused}^{(5)}&=LPF\big(F_{down}^{(4)},I_{\mathrm{NN}}^{4}\big),\end{aligned}
\end{equation}
where $Downsample(\cdot)$ represents the downsampling module, consisting of invPixelShuffle layers and BatchNorm layers.

% \begin{figure*}[ht]
% \centering
% \includegraphics[width=1\linewidth]{SCAB.pdf}
% \caption{\label{fig:SCAB}
% The architecture of the proposed SCAB consists of RCAB and a self-attention mechanism.}
% \end{figure*}

We continue to downsample the features after the \textbf{fifth} fusion until the feature map size is $128 \times 128$, which involves performing two more rounds of feature fusion.
\begin{equation}
\begin{aligned}
{F}_{down}^{(2)} &=Downsample\left(\begin{array}{c}f_{3\times3,64}\left(F_{fused}^{(5)}\right)\end{array}\right),  \\
{F}_{fused}^{(6)} &=\begin{array}{c}LPF(F_{down}^{(2)},{I}_{\mathrm{NN}}^2),\end{array}  \\
{F}_{down}^{(1)} &= Downsample\left(\begin{array}{c}f_{3\times3,64}\left(F_{fused}^{(6)}\right)\end{array}\right),  \\
{F}_{fused}^{(7)} &=\begin{array}{c}LPF(F_{down}^{(1)},{I}_{\mathrm{NN}}^1).\end{array} 
\end{aligned}
\end{equation}

% \begin{equation}
% \begin{aligned}
% {F}_{down}^{(1)} &= Downsample\left(\begin{array}{c}f_{3\times3,64}\left(F_{fused}^{(6)}\right)\end{array}\right),  \\
% {F}_{fused}^{(7)} &=\begin{array}{c}LPF(F_{down}^{(1)},{I}_{\mathrm{NN}}^1).\end{array} 
% \end{aligned}
% \end{equation}

At this point, the feature map, having undergone \textbf{seven} rounds of fusion, contains rich parsing map information and contextual information. Based on this, we employ an Encoder-Decoder strategy to complete the final super-resolution.

In the Encoder part, inspired by\cite{RCAN}, and considering that deeper networks can extract more comprehensive features, we enhanced the feature extraction capability by improving the RCAB (Residual Channel Attention Block) they proposed. We incorporated a convolution-based self-attention module into the RCAB, and we refer to the combined RCAB as SCAB.

Specifically, building upon feature extraction within RCAB, we utilize $1 \times 1$ convolutions to map the input features into three subspaces. Each subspace is further divided into 4 heads. Additionally, we employ depthwise separable convolutions to enlarge the receptive field, thereby encoding channel-level context and generating $Q, K, V \in \mathbb{R}^{C \times H \times W}$.
\begin{equation}
\begin{aligned}
Q&=f_{1\times1}(f_{dconv}^{3\times3}(F)),\\K&=f_{1\times1}(f_{dconv}^{3\times3}(F)),\\V&=f_{1\times1}(f_{dconv}^{3\times3}(F)).\end{aligned}
\end{equation}

By computing the correlation between Q and K, we can obtain global attention weights from different positions, thereby capturing global information. This process can be described as:
\begin{equation}
\begin{aligned}F_\text{att}=\text{Softmax}\left(Q\cdot K/\sqrt{d}\right)\cdotp V,\end{aligned}
\end{equation}
where $\sqrt{d}$ is the scaling factor applied to the dot product result.

For each downsampling, the size of the feature map is halved, and we use two cascaded SCABs to extract further features at depth. In order to make full use of the information of the analytic graph, nearest neighbor downsampling is performed on the analytic graph, and the parsing map is downsampled to $64 \times 64, 32 \times 32, 16 \times 16$, and $8 \times 8$. In the process of each downsampling Encoder, the analytic graph corresponding to the size of the feature graph is fused again.

\begin{equation}
\begin{aligned}\begin{aligned}F_{\textit{feature}}&=LPF\left(\mathrm{SCAB}\left(F_{fused}\right)_{\times2},I_{\mathrm{NN}}^{Down}\right)\\F_{Encoder}&=(Downsample\left(F_{\textit{feature}}\right))_{\times4}.\end{aligned}\end{aligned}
\end{equation}

In the Decoder section, upsampling modules are employed. Similarly, after each upsampling operation, two SCAB modules are concatenated until the feature map size reaches $128 \times 128$ after four upsampling operations.

\begin{equation}
\begin{aligned}
F_{Decoder}=\left(\mathrm{SCAB}\left(Upsample\left(F_{{feature}}\right)\right)_{\times 2}\right)_{\times 4}
\end{aligned}
\end{equation}

\begin{figure*}[ht]
\centering
\includegraphics[width=1\linewidth]{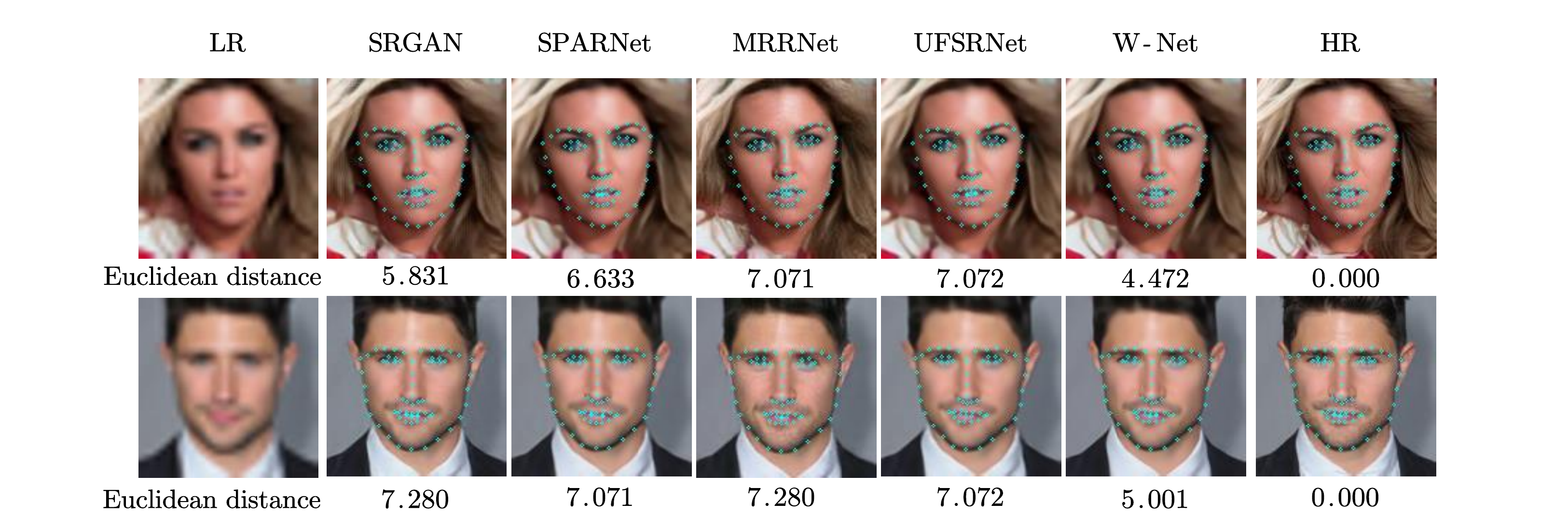}
\caption{\label{fig:keypoint}
Different FSR methods use openface to detect the Euclidean distance between face key points and HR key points.Better zoom in to see the detail.}
\end{figure*}

\begin{figure}[ht]
\centering
\includegraphics[width=1\linewidth]{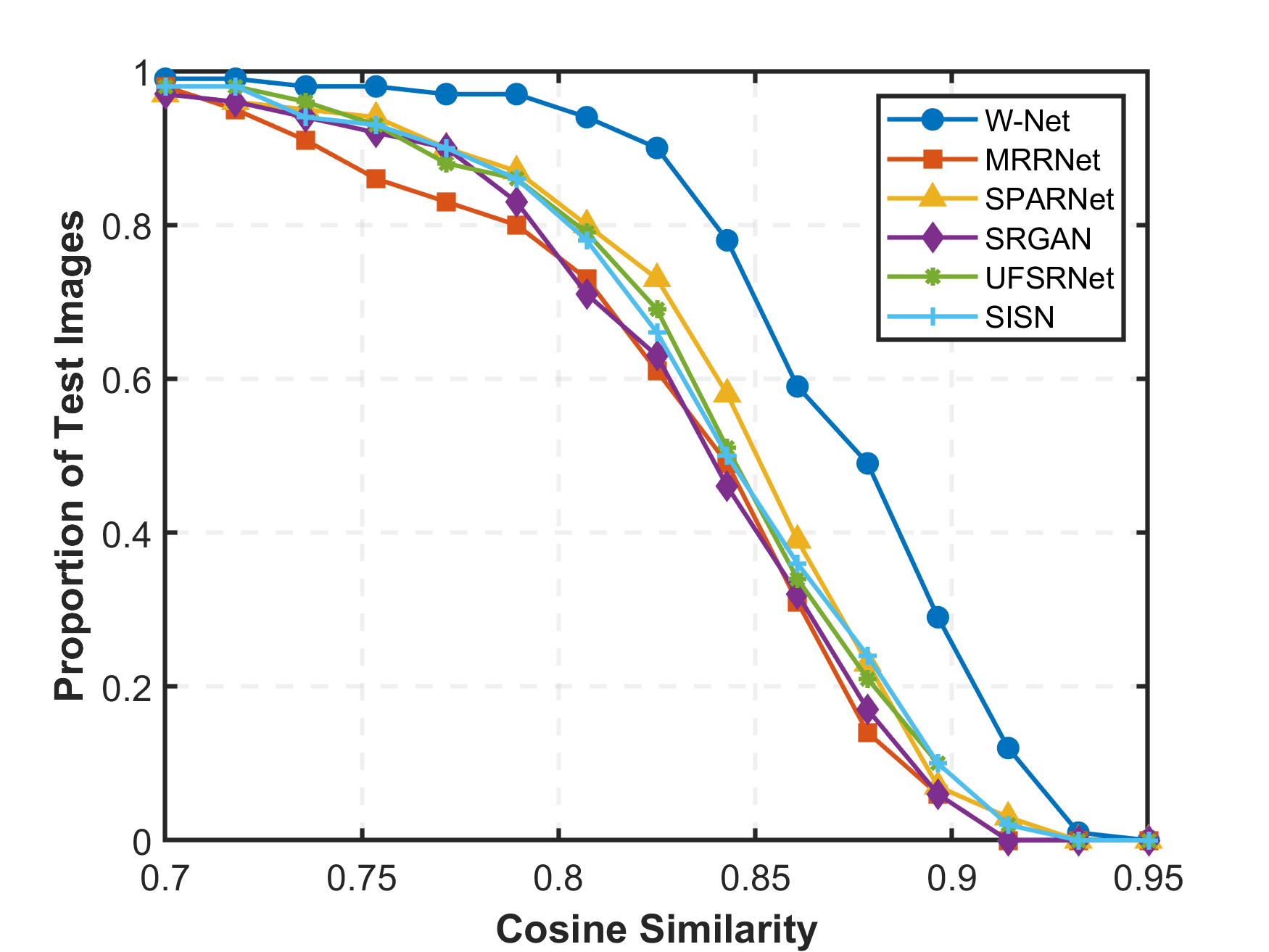}
\caption{\label{fig:Cosine}
With Training Mode II, Identity similarity comparison with other SR models.}
\end{figure}

\subsection{Learning Objectives}
To train our W-Net, two types of loss functions are collaboratively employed: reconstruction loss and perceptual loss\cite{perceptual}. The reconstruction loss is utilized to constrain the approximation of low-resolution images to their corresponding high-resolution ground truth, encompassing both the reconstruction loss of facial parsing maps and the pixel-wise loss of reconstructed facial images. The perceptual loss is then selectively applied with weighting to key facial components.

The pixel-wise loss for facial reconstruction can be defined as:
\begin{equation}
\begin{aligned}
\mathcal{L}_{mse}=\frac{1}{CHW}\|I^{SR}-I^{HR}\|^2.
\end{aligned}
\end{equation}

The pixel-wise loss for facial parsing map reconstruction can be defined as:
\begin{equation}
\begin{aligned}
\mathcal{L}_{parmse}=\frac1{CHW}\|Parsing^{SR}-Parsing^{HR}\|^2.
\end{aligned}
\end{equation}

The perceptual loss is defined as:
\begin{equation}
\begin{aligned}
\mathcal{L}_{per}=\frac{1}{C_kW_kH_k}\parallel\phi_k(I^{HR})-\phi_k(I^{SR})\parallel_2^2,
\end{aligned}
\end{equation}
where $C_k$, $H_k$, and $W_k$ are the dimensions from the k-th
convolution layer of the pretrained VGG-19 model $\phi$\cite{VGG}

The perceptual loss is obtained from facial parsing map 0-1 masks. To better recover information from key facial areas, we employ pixel-wise loss and perceptual loss on four regions (eyes, eyebrows, nose, mouth).
\begin{equation}
\begin{aligned}
\mathcal{L}_{eye}& {=}{L}_{mse}^{eye}{+}{L}_{per}^{eye},  \\
\mathcal{L}_{nose}& {=}\mathcal{L}_{mse}^{nose}{+}\mathcal{L}_{per}^{nose},  \\
\mathcal{L}_{mouth}& {=}\mathcal{L}_{mse}^{mouth}{+}\mathcal{L}_{per}^{mouth},  \\
\mathcal{L}_{eyebrow}& {=}\mathcal{L}_{mse}^{eyebrow}{+}\mathcal{L}_{per}^{eyebrow}. 
\end{aligned}
\end{equation}

We define the keypoint loss as the sum of these four component loss functions:
\begin{equation}
\begin{aligned}
\mathcal{L}_{key}=\mathcal{L}_{mouth}+\mathcal{L}_{nose}+\mathcal{L}_{eye}+\mathcal{L}_{eyebrow}.
\end{aligned}
\end{equation}
In summary, the overall loss of the Fine-grained W-Net is:
\begin{equation}
\begin{aligned}
\mathcal{L}_{fine}=\lambda_{pixel}\mathcal{L}_{mse}+\lambda_{par}\mathcal{L}_{parmse}+\lambda_{key}\mathcal{L}_{key},
\end{aligned}
\end{equation}
where $\lambda_{pixel},\lambda_{par},\lambda_{key}$ represents the weighting coefficient for the loss functions, and the eyes, nose, eyebrows, and mouth are all considered key facial components with equal weighting. The reconstruction pixel-wise MSE loss and the facial parsing MSE loss each have a weighting of $\lambda_{pixel},\lambda_{par}$.

\section{Experiment}
\subsection{Datasets and Metrics}
In this paper, experiments are conducted on two widely used datasets.

\textbf{CelebA:\cite{CelebA}}CelebA consists of a large and diverse set of subjects, exhibiting diversity in poses and categories, comprising over 200,000 images. Our W-Net training data is sourced from this dataset. To comprehensively evaluate the practicality of the model, we follow the processing approaches of DIC\cite{DIC} and RAAN\cite{RAAN}, selecting 168,854 images (Training Mode I) and 18,000 images (Training Mode II) respectively for W-Net training. For Training Mode I, 100 images are used for validation, and 1000 images are used for testing. For Training Mode II, 100 images are used for testing. It's worth noting that the test set is neither used for training nor for validation purposes.

\textbf{Helen\cite{Helen}}: Helen comprises 2330 facial images with 194 facial landmarks. Following the experimental setup of DIC \cite{DIC}, we only utilize 50 images for testing the models trained using Training Mode I and Training Mode II, respectively.

We utilize three commonly used metrics to evaluate the super-resolution (SR) results, including Peak Signal-to-Noise Ratio (PSNR), Structural Similarity (SSIM)\cite{SSIM}, and Learned Perceptual Image Patch Similarity (LPIPS)\cite{LIPIS}. A higher PSNR and SSIM indicate smaller differences between two images. Conversely, a smaller LPIPS indicates greater similarity between two images.

\subsection{Implementation Details}
In CelebA, the facial images have inconsistent heights and widths. Therefore, we employ OpenFace\cite{openface} to detect 68 facial keypoints and crop the images based on these keypoints, resizing them to $128 \times 128$ pixels as HR. Subsequently, we further downsample the HR images using bicubic interpolation to $32 \times 32 $ and $16 \times 16$, serving as LR facial images for $\times 4$ and $\times8$ FSR, respectively. For the parsing map dataset, we adopt a pre-trained BiSeNet\cite{BiseNet} for facial segmentation, followed by a binarization operation on the segmented maps to simplify the parsing map problem, obtaining high-resolution parsing map images. Similarly, we downsample the HR parsing maps using bicubic interpolation to $32 \times 32 $ and $16 \times 16$, serving as LR parsing map images for $\times 4$ and $\times 8$ FSR ground truth. We utilize the popular Adam optimizer ($\beta_1=0.90,\beta_2=0.999,\epsilon =1e-8$), and employ the loss weight $\lambda_{pixel}=1.0,\lambda_{par}=1.0,\lambda_{key}=0.5$ during the training of W-Net. The learning rate is set to $1e-4$. Our experiments are implemented on PyTorch\cite{Pytorch} using four NVIDIA 4090 GPUs. Generally, with Training Mode I and a batch size of 4, training a W-Net takes 45 hours; with Training Mode II and a batch size of 4, training a W-Net takes 9 hours.

\subsection{Comparisons With State-of-the-Arts}
To demonstrate the quantitative advantages of our W-Net, we compare our method with state-of-the-art FSR and general SR methods under $4 \times$ and $8 \times$ downsampling factors, both qualitatively and quantitatively. The comparison methods include three representative image super-resolution methods: SRCNN\cite{SRCNN}, EDSR\cite{EDSR}, and VDSR\cite{VDSR}; two methods that do not use any prior knowledge: MRRNet\cite{MRRNet} and SRGAN\cite{SRGAN}; methods that also use prior knowledge: SPARNet\cite{SPARNet} and DIC\cite{DIC}; methods using deep attention networks: RAAN\cite{RAAN} and SISN\cite{lu2021face}; a method leveraging wavelet transforms: WIPA\cite{WIPA}; and the recently proposed UFSRNet\cite{UFSRNet}. To ensure fairness, all comparison models were retrained on the same dataset (Training Mode I with 168,854 images and Training Mode II with 18,000 images, all from CelebA). Table ~\ref{tab:Compare}show the PSNR and SSIM results on the Helen and CelebA datasets under training modes I and II, respectively. It can be observed that our W-Net outperforms other methods in terms of PSNR and SSIM metrics.

The experimental results of $4 \times$ and $8 \times$ upscaling under Training Mode II are shown in Figure ~\ref{fig:Compare}. From the subjective visual quality perspective, our method demonstrates the best performance. While other compared FSR algorithms also effectively reconstruct facial images, they tend to overly smooth facial details. For instance, the eye regions of the two individuals reconstructed by MRRNet and UFSRNet are excessively blurry. Based on our analysis, this is likely due to their insufficient use of prior information, leading to discrepancies between the reconstructed facial images and the real ones. For SPARNet, the loss of information during the feature extraction process results in overly smooth details, failing to restore specific details like double eyelids. As for WIPA and SISN, many attribute information of the face is lost and the eye area is more blurred, which we believe is caused by its failure to utilize prior knowledge. On the other hand, SRGAN performs better in restoring eye details compared to other models but at the expense of pixel accuracy. Our method achieves a balance between the perceptual quality and pixel accuracy of super-resolved facial images by fully leveraging the information from parsing maps and integrating a comprehensive loss function for key areas. We also used LPIPS to evaluate our method and other methods. LPIPS reflects perceptual similarity based on deep features. MRRNet focuses on perceptual similarity, thus having an advantage in LPIPS metrics, but its PSNR and SSIM are significantly affected. In contrast, our W-Net ensures high PSNR and SSIM while also performing well in LPIPS. Although there is a slight gap in LPIPS compared to MRRNet, our W-Net achieves a better overall balance.

Additionally, we conducted comparative experiments on downstream tasks. First, we used the pre-trained facial recognition model AdaFace\cite{adaface} to extract identity feature vectors from both SR and HR. We then calculated the cosine similarity between these feature vectors to measure identity similarity between the SR and HR faces. Figure ~\ref{fig:Cosine} shows the proportion of test images with a cosine similarity above a specific threshold. The results indicate that our proposed model preserves identity better in super-resolved images than other SR models.

Moreover, we used the pre-trained OpenFace\cite{openface} model to detect 68 landmarks on both SR and HR images and calculated the Euclidean distance between the landmark results of SR and HR images. The visualized comparison results are shown in the Figure ~\ref{fig:keypoint}. The quantitative results of these two comparative experiments are shown in Table ~\ref{tab:xiayou}, demonstrating that our proposed model has the greatest advantage in downstream tasks.

\begin{figure*}[htb!]
    \centering
    \includegraphics[width=1\linewidth]{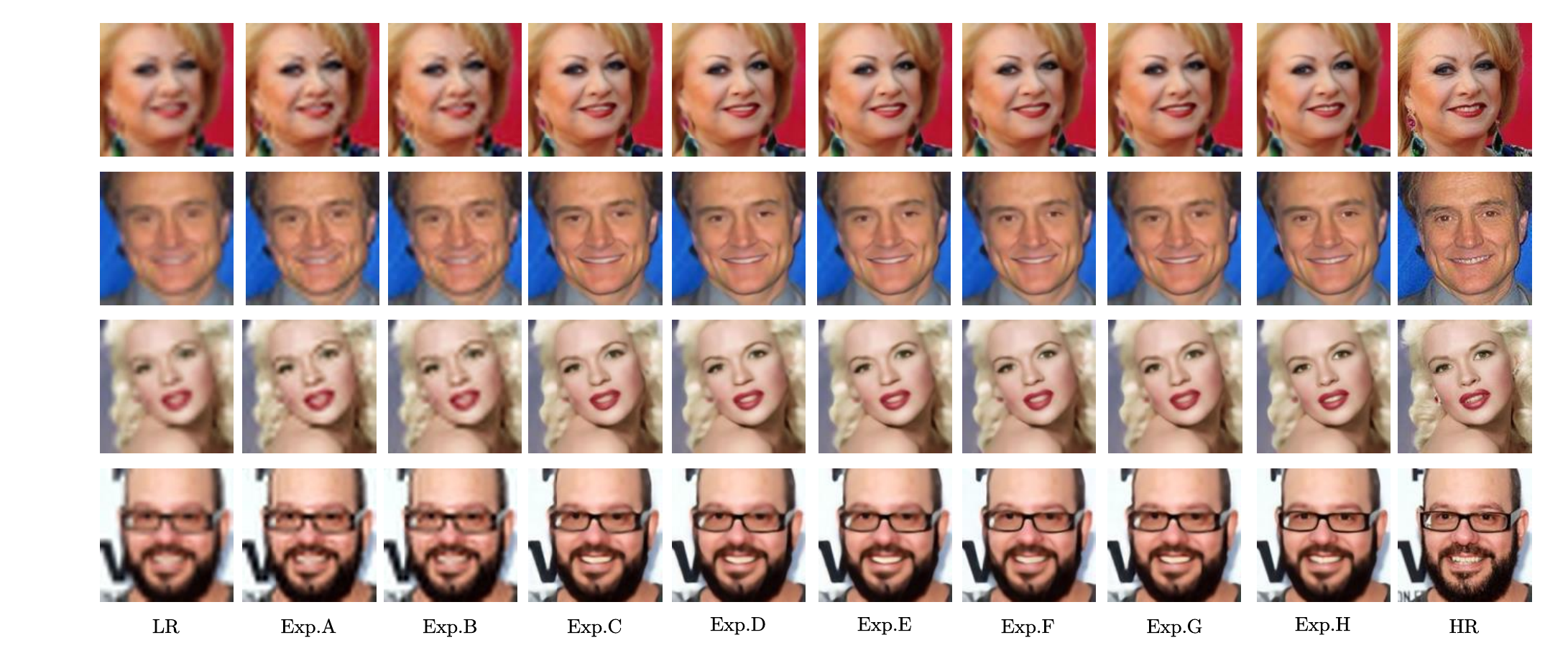}
    \caption{Visual differences in ablation tests. Better zoom in to see the detail.}
    \label{fig:xiaorong}
\end{figure*}

\begin{figure}[ht]
\centering
\includegraphics[width=1\linewidth]{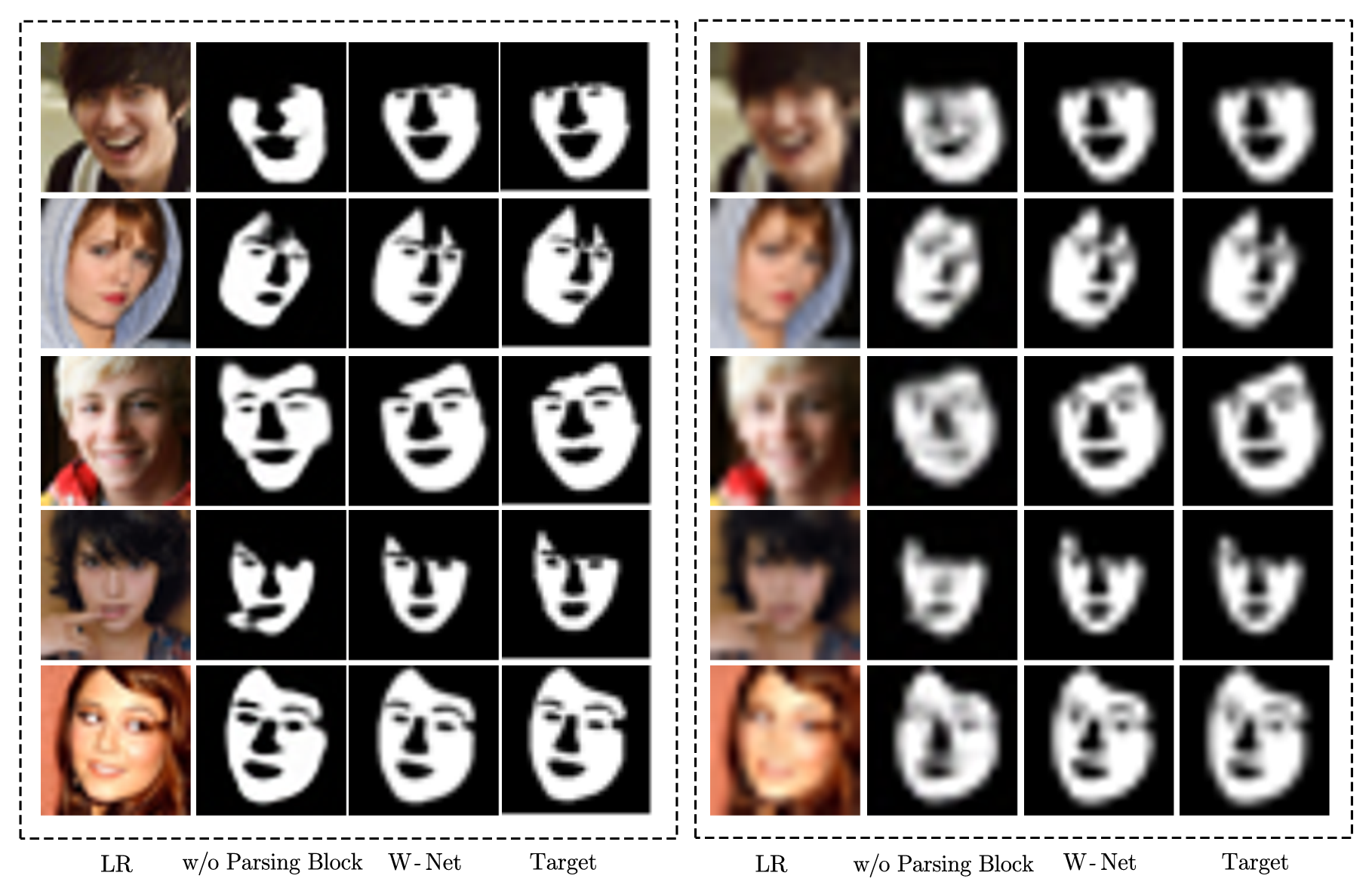}
\caption{\label{fig:jiexitu}
Visualization of the influence of the parsing module on generating parsing maps: on the left are the parsing map comparisons at a $\times 4$ scale, and on the right are the parsing map comparisons at a $\times 8$ scale. Better zoom in to see the detail.}
\end{figure}

\subsection{Ablation Study}
In this section, we analyze the effectiveness of the proposed Parsing Block, LPF, and SCAB. For a clear comparison, we modified LPF to a basic cascade, replaced the Parsing Block with a standard ConvBlock, and substituted SCAB with the original RCAB module. We refer to this modified base model as ExpA. Based on this model, we conducted a series of experiments, with the results presented in Table ~\ref{tab:xiaorong}.

\textbf{The Effectiveness of Parsing Block:} 
To validate the effectiveness of the proposed Parsing Block, we replaced the ordinary convolution in Exp.A with the Parsing Block to create Exp.B. It can be observed from Table ~\ref{tab:xiaorong} that the model performance of Exp.B is slightly better than that of Exp.A, but the improvement is not significant. We believe that although the Parsing Block brings better parsing maps, the lack of utilization of the parsing map information due to the absence of LPF may have contributed to this result. Hence, we simultaneously employed LPF and Parsing Block as Exp.F, resulting in a significant improvement in PSNR.

To further illustrate the function of the Parsing Block, we conducted comparative experiments on the parsing images with relatively low quality($\times 4$) and extremely low quality($\times 8$), as shown in Figure ~\ref{fig:jiexitu}. After integrating the Parsing Block, our model demonstrates improved extraction of parsing image features, such as the nose and mouth positions. The parsing images generated by our model are noticeably closer to the target images.

\textbf{The Effectiveness of LPF:} In this series of experiments, we aimed to validate the effectiveness of LPF. We replaced the multiple fusions of LPF in the W-Net with a basic cascade, denoted as Exp.A. Then, we substituted LPF into the W-Net, labeled as Exp.C. Furthermore, we combined LPF with SCAB in Exp.F. We found that the LPF module significantly improves the PSNR value. We attribute this improvement to LPF's ability to effectively fuse information from low-quality images and parsing maps, thereby balancing the prior knowledge of facial structure. As shown in Table ~\ref{tab:xiaorong}, Exp.F and Exp.G with LPF respectively achieved the second-best PSNR and SSIM metrics.

\begin{table}[h]
    \centering
    % \small
    \footnotesize
    \caption{Analysis of the effectiveness of different modules of the proposed model on the CelebA dataset, using Training Mode II and a scale factor of $\times 4$.}
    \begin{tabular}{cccccc}
        \toprule
        Methods & Parsing Block & LPF & SCAB & PSNR & SSIM \\
        \midrule
        Exp.A &  &  &  & 28.69 & 0.8358 \\
        Exp.B & \checkmark &  &  & 28.98 & 0.8438 \\
        Exp.C &  & \checkmark &  & 31.01 & 0.8903 \\
        Exp.D &  &  & \checkmark & 31.10 & 0.8956 \\
        Exp.E & \checkmark &  & \checkmark & 31.21 & 0.8965 \\
        Exp.F &  & \checkmark & \checkmark & 31.36 & \underline{0.8977} \\
        Exp.G & \checkmark & \checkmark &  & \underline{31.37} & 0.8973 \\
        Exp.H & \checkmark & \checkmark & \checkmark & \textbf{31.63} & \textbf{0.9029} \\
        \bottomrule
    \end{tabular}
    \label{tab:xiaorong}
\end{table}

\textbf{The Effectiveness of SCAB:} We added SCAB after each upsampling and downsampling layer as a feature extraction method. The model using SCAB is referred to as Exp.B. With the addition of the self-attention mechanism, the aim is to explore the relationship between global and local features. It achieves a better balance between these features, resulting in an improvement in PSNR  compared to RCAB.

Simultaneously using three modules, namely the W-Net adopted in this paper, allows us to achieve the performance of our final trained model. Through these experiments, we can conclude that in FSR, the complementary multiple integrations between the face parsing maps and the low-quality images play an important role.

\section{Conclusion}
This paper proposes a W-Net for FSR , which utilizes facial parsing maps from facial structure priors during reconstruction. We designed a Parsing Block to exploit the potential of obtaining facial parsing maps from low-quality images. Based on this, we developed an LPF  module to integrate the information from parsing maps and low-quality images. Thanks to our W-shaped network architecture, the LPF module is used multiple times across various dimensions, allowing us to obtain richer features. Moreover, to balance perceptual quality and pixel accuracy, we use facial parsing map masks to assign different weights and loss functions to key facial areas. Finally, we conducted extensive comparative experiments to validate the feasibility of the proposed method. Our model demonstrated advantages in technical metrics and pixel accuracy, providing a promising direction for future FSR research.

\section{Acknowledgments}
This research was funded by the Shandong Provincial Natural Science Foundation of China, grant numbers ZR2020MF027 and ZR2020MF143. 
{\small
\bibliographystyle{unsrt}
\bibliography{main}
}

\end{document}